%% file: dmf.tex
\def\year{2018}\relax
\documentclass[letterpaper]{article} 
\usepackage{aaai18}  
\usepackage{times}  
\usepackage{helvet}  
\usepackage{courier}  
\usepackage{array}
\usepackage{mathtools}
\usepackage{enumitem}
\usepackage{dsfont}
\usepackage{url}  
\usepackage{graphicx, subfigure}
\usepackage{amssymb}  
\usepackage[linesnumbered,boxed,ruled]{algorithm2e}
\begin{document}
%
\title{Privacy Preserving Point-of-interest Recommendation Using Decentralized \\Matrix Factorization}
\author{Chaochao Chen, Ziqi Liu, Peilin Zhao, Jun Zhou, Xiaolong Li\\
~\\AI Department, Ant Financial Services Group\\
~\\\{chaochao.ccc,ziqiliu,peilin.zpl,jun.zhoujun,xl.li\}@antfin.com\\
}
\maketitle
\begin{abstract}
Points of interest (POI) recommendation has been drawn much attention recently due to the increasing popularity
of location-based networks, e.g., Foursquare and Yelp. 
Among the existing approaches to POI recommendation, Matrix Factorization (MF) based techniques have proven to be effective. 
However, existing MF approaches suffer from two major problems:
(1) Expensive computations and storages due to the centralized model training mechanism: the centralized
learners have to maintain the whole user-item rating matrix, and potentially huge low rank matrices.
(2) Privacy issues: the users' preferences are at risk of leaking to malicious attackers via
the centralized learner.
To solve these, we present a Decentralized MF (DMF) framework for POI recommendation. 
Specifically, instead of maintaining all the low rank matrices and sensitive rating data for training, we propose a
random walk based decentralized training technique to train MF models on each user's end, e.g., cell phone and Pad. 
By doing so, the ratings of each user are still kept on one's own hand, and moreover, decentralized learning can
be taken as distributed learning with multi-learners (users), and thus alleviates the computation and storage issue. 
Experimental results on two real-world datasets demonstrate that, comparing with the classic and state-of-the-art latent factor models, DMF significantly improvements the recommendation performance in terms of precision and recall.
\end{abstract}

\input{section/intro}

\input{section/relatedwork}

\input{section/model} 
\input{section/experiment}

\input{section/conclusion}

\bibliographystyle{aaai}
\bibliography{dmf-reference}

\end{document}

%% file: section/intro.tex
\section{Introduction}

Nowadays, location-based networks, e.g., Foursquare and Yelp, are becoming more and more popular. 
These platforms provide kinds of point of interests (POIs) such as hotels, restaurants, and markets, which makes our lives much easier than before. 
Meanwhile, the problem of ``where to go'' starts to bother people, since it is time-consuming for people to find their desired places from so many POIs. 
POI recommendation appears to address such a problem by helping users filter out uninteresting POIs and saving their decision-making time \cite{ye2011exploiting,gao2015content}.

Among the existing research from POI recommendation communities, Matrix Factorization (MF) techniques draws a lot of attention \cite{cheng2011exploring,cheng2012fused,yang2013sentiment}, and it has proven to be attractive in many recommendation applications \cite{koren2009matrix,chen2014context,koenigstein2011yahoo,chen2016capturing}. 
However, existing MF approaches train the recommendation model in a centralized way. 
That is, the recommender system, actually the organization who built it, first needs to collect the action data of all the users on all the POIs, and then trains a MF model. 
The shortcomings of this kind of centralized MF methods are mainly three-folds.
(1) \textit{High cost of resources}. 
Centralized MF not only needs large storage to store the collected user action data on POIs, but also requires huge computing resources to train the model, especially when datasets are large. 
(2) \textit{Low model training efficiency}. Centralized MF model is trained on a single machine or a cluster of machines. 
Thus, the model training efficiency is restricted to the number of machines available. 
(3) \textit{Shallow user privacy protection}. 
In recommender systems, user privacy is an importance concern \cite{canny2002collaborative}, especially in POI scenarios. 
Most people do not want to release their activities on location based network, not only to other people, but also to the platforms, for privacy concerns. 
However, centralized MF trains model on the basis of having all the users' activities data. 
In fact, the safest way is that users' data are kept in their own hand, without sending them to anyone or any organization (platform).

\begin{figure}[!tp]
\centering
\includegraphics[width=7cm]{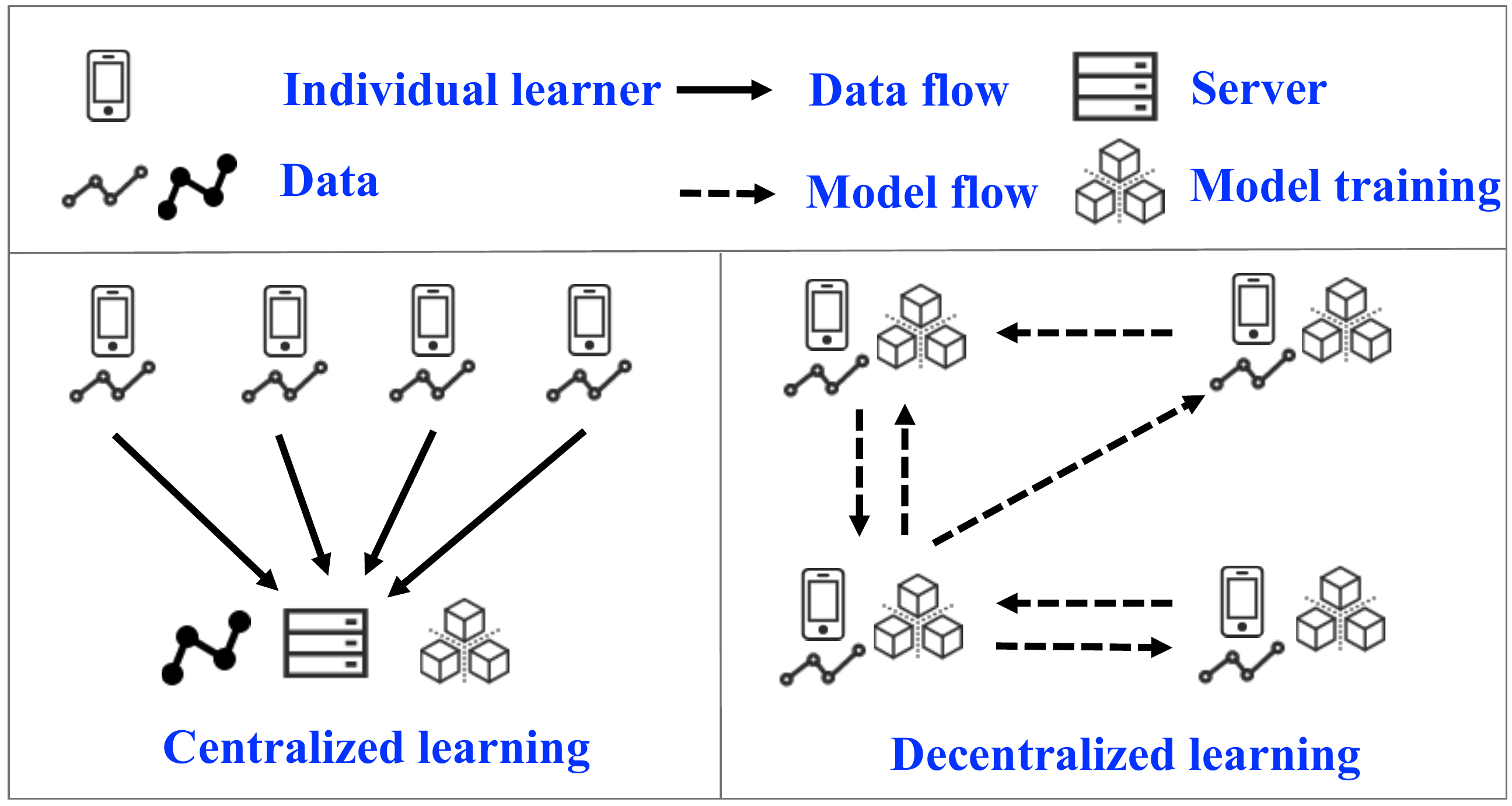}
\vskip -0.1in
\caption{Comparison between centralized learning and decentralized learning.}
\label{framework}
\vskip -0.15in
\end{figure}
 
To solve the above problems caused by centralized training of MF, we propose Decentralized MF (DMF) framework. 
That is, instead of training user and item latent factors in a centralized way by obtaining all the ratings, we train MF model in each user's end, e.g., user's cell phone and Pad.  
Figure \ref{framework} shows the difference between centralized and decentralized learning. 
DMF treats each user's device as an autonomous learner (individual computation unit). 
As we know, the essence of MF is that user and item latent factors are learnt collaboratively, so should be DMF. 
Three key challenges exist when making users collaborate in DMF meanwhile preserving user privacy. 
The first challenge is \textit{which user should be communicated}. 
We answer this question by analyzing the data in POI recommendation scenarios, and propose nearby user communication on user adjacent graph, which is built based on users' geographical information. 
Along with this, the second challenge naturally appears: \textit{how far should users communicate with their neighbors}. 
To address this, we present a random walk method to enhance local user communication. 
That is, we use random walk technique to intelligently select users' high-order neighbors instead of only their direct neighbors. 
The third and also the most serious challenge is \textit{what information should users communicate with each other} without leaking their data privacy. 
To solve this challenge, we decompose item preference into global (common) and local (personal) latent factors, and allow users communicate with each other by sending the gradients of the global item latent factors. 

Our proposed DMF framework successfully deals with the shortcomings of centralized MF.  
(1) Each user only needs to store his own latent factor, and items' latent factors. 
The computation is also cheap: each user only need to update the corresponding user and item latent factors when he (his neighbors) rates (rate) an item. 
(2) Since each user updates the DMF model on his own side, it can be taken as a distributed learning system with \#users as \#machines, which makes the model efficient to train. 
(3) The ratings of each user on items are still kept on one's own hand, which avoids user's privacy being disclosed.

We summarize our main contributions as follows: 
\begin{itemize}
\item We propose a novel DMF framework for POI recommendation, which is scalable and is able to preserve user privacy. 
To our best knowledge, it is the first attempt in literature. 
\item We propose an efficient way to train DMF. Specifically, when a user rates an item on his side, the user and item gradients are first calculated. We then present a random walk based technique for users to send the item gradient to their neighbors. Finally, the correspoding user/item latent factors are updated using stochastic gradient descent. 
\item Experimental results conducted on two real-world datasets demonstrate that DMF can achieve even better performance compared with the classic and state-of-the-art latent factor models in terms of precision and recall. 
Parameter analysis also shows the effectiveness of our proposed random walk based optimization method. 
\end{itemize}

%% file: section/relatedwork.tex
\section{Background}\label{background}
In this section, we review some necessary backgrounds which form the basis of our work, i.e., (1) Matrix Factorization (MF) in POI Recommendation, (2) decentralized learning.

\subsection{Matrix Factorization in POI Recommendation}

MF aims to learn user and item (POI) latent factors through regressing over the existing user-item ratings \cite{koren2009matrix,mnih2007probabilistic}, which can be formalized as follows,
\begin{equation}\small
\label{lfm}
\begin{split}
\mathop {\arg \min }\limits_{u_i,v_j} \sum\limits_{i=1}^{I} \sum\limits_{j=1}^{J} {{\left( {r_{ij} - u_i^T{v_j}} \right)}^2}  + \lambda\left(\sum\limits_{i=1}^{I} {||{u_i}||^2}  + \sum\limits_{j=1}^{J} {||{v_j}||^2}\right),
\end{split}	
\end{equation}
where $u_i$ and $v_j$ denote the latent factors of user $i$ and item $j$, respectively, and $r_{ij}$ denotes the known rating of user $i$ on item $j$. We will describe other parameters in details later. 

MF and its variants have been extensively applied to POI recommendation due to their promising performance and scalability \cite{cheng2011exploring,cheng2012fused,yang2013sentiment}. 
However, these methods are all trained by using the centralized mechanism. 
This centralized MF training results in expensive resources required, low model training efficiency, and shallow protection of user privacy.

\subsection{Decentralized Learning}
Decentralized learning appears to solve the above problems of centralized learning 
\cite{nedic2009distributed,yan2013distributed}. 
Recently, it has been applied in many scenarios such as multiarmed bandit \cite{kalathil2014decentralized}, network distance prediction \cite{liao2010network}, hash function learning \cite{leng2015hashing}, and deep networks \cite{shokri2015privacy,mcmahan2016communication}. 

The most similar existing works to ours are decentralized matrix completion \cite{ling2012decentralized,yun2014nomad}. 
However, we summarize the following two major differences. 
(1) They either allow each learner (user) to communicate with those who have rated the same items or communicate with all the learners, and thus have low accuracy or high communication cost. 
In practice, users always collaborate with their affinitive users. 
To capture this, in this paper, we propose a random walk approach for users from the adjacent graph to collaboratively commununicate with each other. 
(2) They allow directly exchange of item preference among learners, which may cause information leakage. 
For example, it is easy to be hacked by using the idea of collaborative filtering \cite{sarwar2001item}, i.e., similar items tend to be preferred by similar users. 
Assume user $i$ is a malicious user, he has his own latent factor of item $j$ ($v^i_j$). 
He also gets the latent factor of item $j$ from user $i'$ ($v^{i'}_j$). 
If $i$ likes item $j$, and $v^i_j$ and $v^{i'}_j$ are similar, then $i$ will know $i'$ likes $j$ as well.
In contrast, in this paper, we propose a gradient exchange scheme to limit the possibility of privacy leakage.

%% file: section/model.tex
\section{The Proposed Model}\label{model}
In this section, we first formally describe the Decentralized Matrix Factorization (DMF) problem. 
We then discuss a nearby user communication scheme for users to collaborate with each other. 
Next, we propose an enhanced version by applying random walk theory. 
Then, we present a privacy preserving nearby user collaboration algorithm to optimize DMF. 
We analyze the model complexity in the end.

\subsection{Preliminary}

Formally, let $\mathcal{U}$ and $\mathcal{V}$ be the user and item (POI) set with $I$ and $J$ denoting user size and item size, respectively. 
Let $(i,j)$ be an interaction between user $i \in \mathcal{U}$ and item $j \in \mathcal{V}$, and $r_{ij}$ be the rating of user $i$ on item $j$. 
Without loss of generality, we assume $r_{ij} \in [0,1]$ in this paper. 
Let $\mathcal{O}$ be the training dataset, where all the user-item ratings in it are known. 

For the traditional centralized MF, it first collects all the $r_{ij} \in \mathcal{O}$, and then learns $\textbf{U}\in \mathbb{R}^{K\times{}I}$ and $\textbf{V}\in \mathbb{R}^{K\times{}J}$ using MF technique (Equation \ref{lfm}). 
Here, $\textbf{U}\in \mathbb{R}^{K\times{}I}$ and $\textbf{V}\in \mathbb{R}^{K\times{}J}$ denote the user and item latent factor matrices, with their column vectors $u_i$ and $v_j$ be the $K$-dimensional latent factors for user $i$ and item $j$, respectively.

For DMF, to guarantee the privacy of each user, we need to keep all the known ratings and latent factors on each user's end during the whole training procedure. 
To do this, we use $\textbf{U}\in \mathbb{R}^{K\times{}I}$ to denote user latent factor matrix, with each column vector $u_i$ denotes the $K$-dimensional latent factors for user $i$. 
We also use $\textbf{V}\in \mathbb{R}^{I\times{}K\times{}J}$ to denote item latent factor tensor, with $\textbf{V}^i\in \mathbb{R}^{K\times{}J}$ denotes the item latent factor matrix for user $i$, and further with $v^i_j$ denotes the $K$-dimensional latent factors for item $j$ of user $i$. 
Thus, each user $i$ only needs to store $i$'s own $K$-dimensional latent factor $u_i$, and $i$'s item latent factor matrix $\textbf{V}^i$.

Besides, users need to collaboratively learn their stored factors, i.e., $u_i$ and $\textbf{V}^i$, in DMF scenario. 
For centralized MF, all the users share the same item latent factor matrix, i.e., $\textbf{V}^i=\textbf{V}^{i'}, \forall ~ i,i' \in \mathcal{U}$. 
For DMF, each user stores his $u_i$ and $\textbf{V}^i$, and they should be trained collaboratively with other users---which we call `neighbors'. 
Suppose we have a user adjacent graph $\mathcal{G}$, we use $\textbf{W}\in \mathbb{R}^{I\times{}I}$ to denote user adjacency matrix, where each element $w_{i,i'} \in [0, 1]$ denotes the degree of relationship between user $i$ and $i'$. Of course, user $i$ and $i'$ have no relationship if $w_{i,i'}=0$. 
We use $\mathcal{N}^d(i)$ to denote the $d^{st}$ order neighbors of $i$ on $\mathcal{G}$, $|\mathcal{N}^d(i)|$ as the neighbor size, and $|\mathcal{N}^D(i)|=\sum\limits_{d}^D|\mathcal{N}^d(i)|$. 
Obviously, $\mathcal{N}^1(i)$ denotes the direct neighbors of $i$. 
Besides, to save communication cost, we use $N$ to denote the maximum number of direct neighbors of each user. 

DMF aims to learn $u_i$ and $\textbf{V}^i$ for each user, and the model learning procedure is performed on one's own side, e.g., cell phone and Pad.

\subsection{Nearby User Communication}
The essence of MF is that the user and item latent factors are learnt collaboratively, so should be DMF. 
Thus, \textit{which user should be communicated under DMF framework} becomes the first challenging question. 
We answer this question by first analyze the data in POI recommendation scenarios. 

\begin{figure}
\centering
\subfigure[Foursquare dataset] { \includegraphics[width=4cm,height=3.3cm]{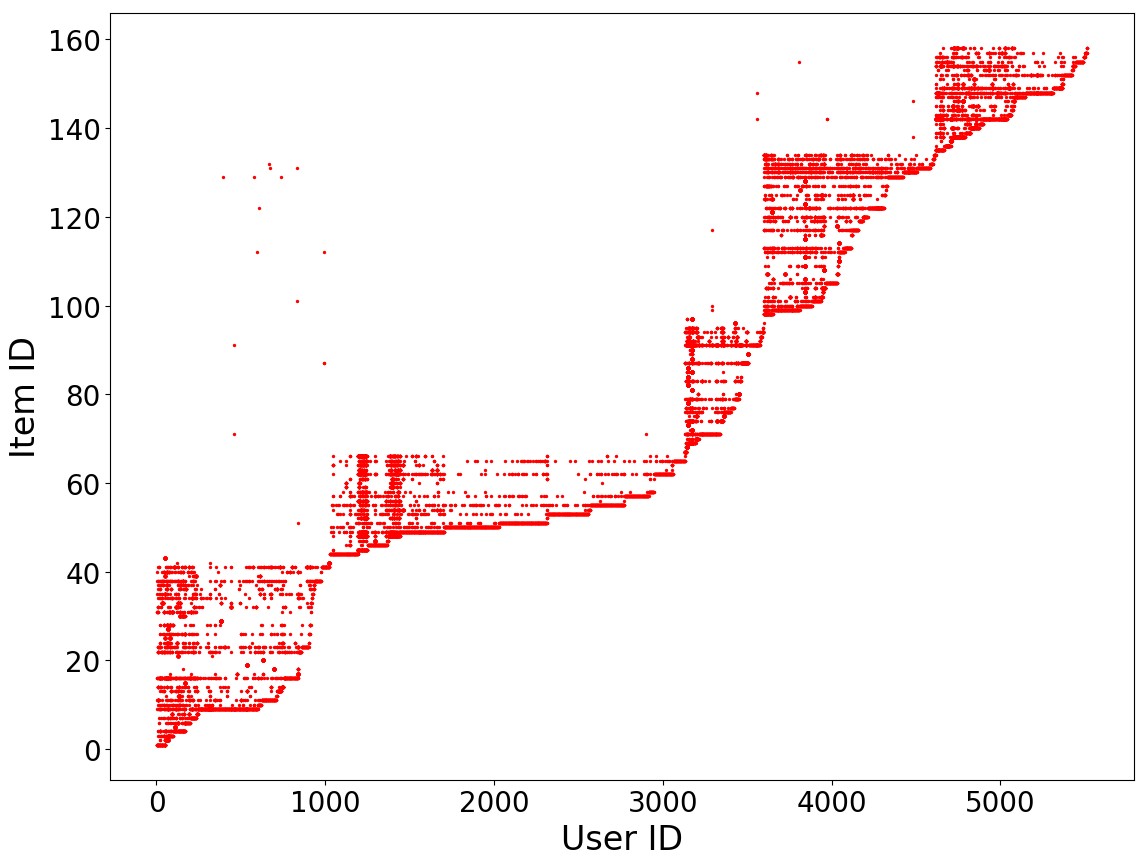}}~~~
\subfigure [Alipay dataset]{ \includegraphics[width=4cm,height=3.3cm]{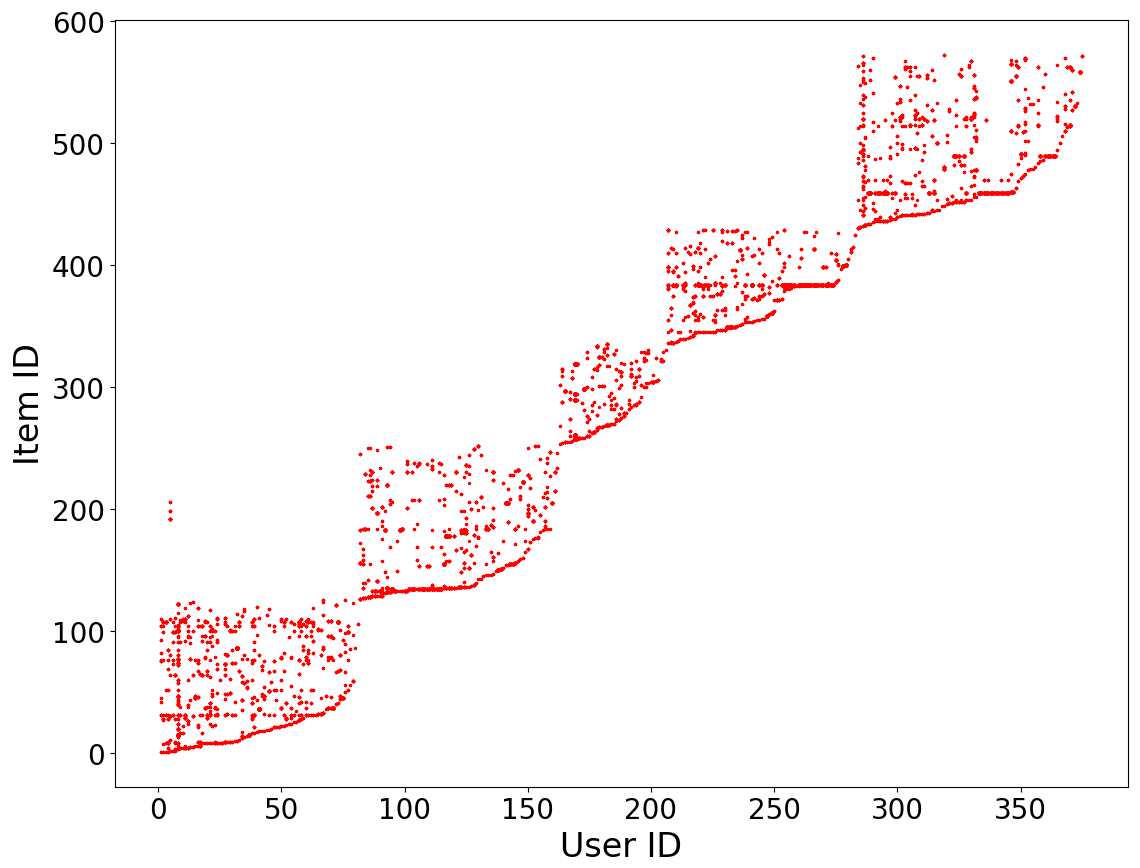}}
\vskip -0.1in
\caption{Data analysis of Foursquare and Alipay datasets.}
\label{dataanalysis}
\vskip -0.1in
\end{figure}

\textbf{Observation.} 
Figure \ref{dataanalysis} shows the user-POI check-in distributions on two real datasets, i.e., \emph{Foursquare} and \emph{Alipay}. 
Both datasets contain user-item-check-in-location records, and we divide locations into different cities. 
We randomly select the user-item check-in records in five cities from both datasets, and plot their check-in distributions in Figure \ref{dataanalysis}, where each dot denotes a user-item check-in record. 
From it, we have the following observation: 
in POI scenarios, most users are only active in a certain city, and we call it ``location aggregation''. 
Only a few users be active in multi-cities, which is neglectable. 


\textbf{User Adjacent Graph. } 
We represent the affinities among users based on the definition of a user adjacency graph. 
It can be built using whatever information that is available, e.g., rating similarity \cite{su2009survey} and user social relationship \cite{yang2011like}. 
However, in DMF for POI recommendation scenario, users' ratings are on their own hand, and social relationship is not always available. 
Thus, we focus on using user geographical information to build user adjacent graph, similar as the existing researches \cite{ye2011exploiting,cho2011friendship,cheng2012fused}. 
Specifically, suppose 
$d_{i,i'}$ is the distance between user $i$ and $i'$, 
the relationship degree between $i$ and $i'$ is defined as 
\begin{equation}\small
w_{i,i'} = I^{i,i'} \Large \cdot f(d_{i,i'}),
\end{equation} 
where $w_{i,i'} \in [0,1]$, $I^{i,i'}$ is the indicator function that equals to 1 if $i$ and $i'$ are in the same city and 0 otherwise, and $f(d_{i,i'})$ is a mapping function of distance and relationship degree, and the smaller the distance of $i$ and $i'$ is, the bigger their relationship degree is. 
Existing research has proposed different such mapping functions \cite{zhao2016survey}. 
In practice, it's extremely expensive if we maintain the communications for those super-users who have a huge number of neighbors. Thus, we set $N$ to be the maximum number of neighbors each user can have. 
With nearby user communication schema, the first question is answered. 

\subsection{Random Walk Enhanced Nearby User Communication}

\begin{figure}[!tp]
\centering
\includegraphics[width=5.5cm]{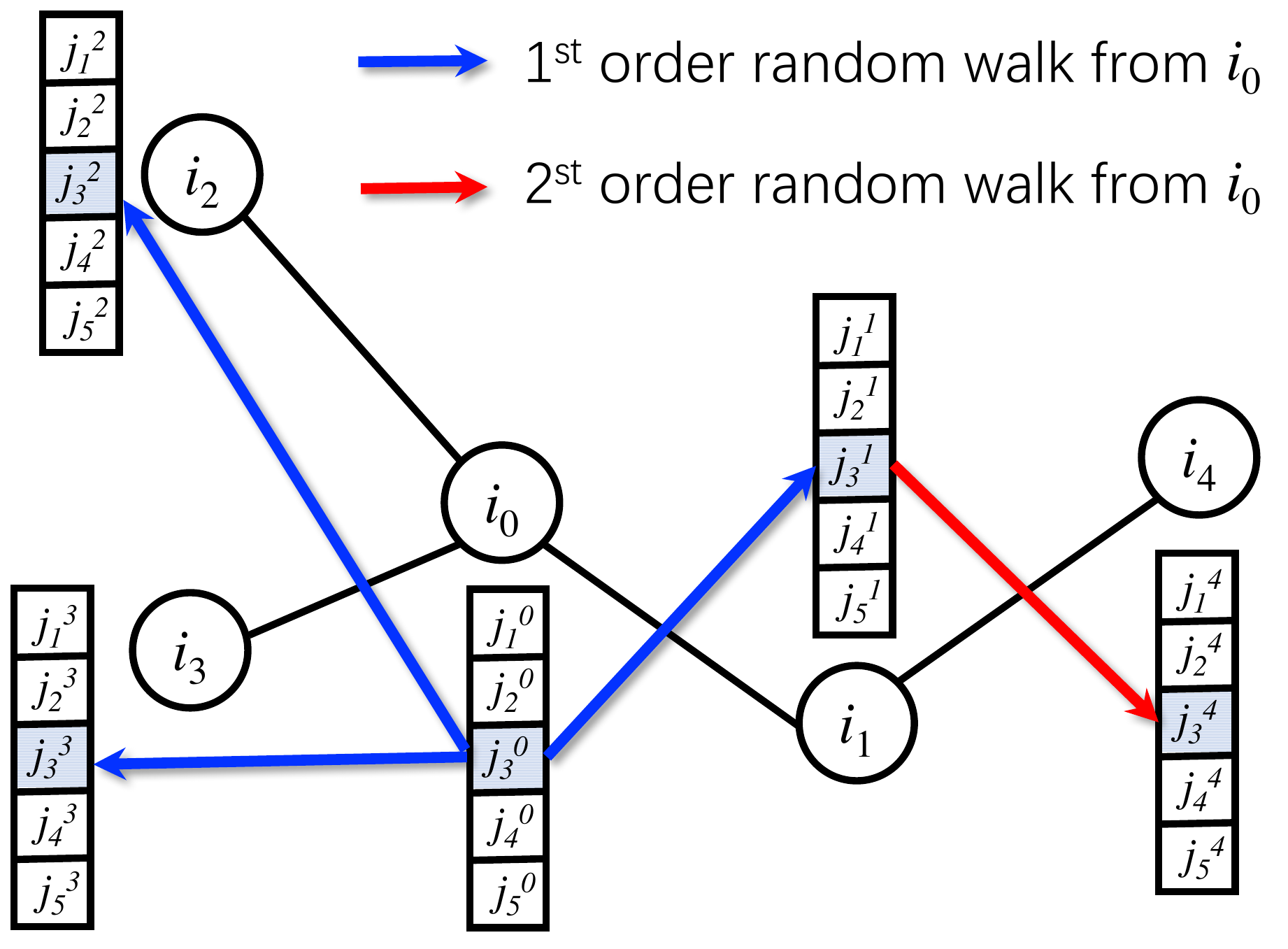}
\caption{Random walk enhanced nearby user communication. User $i_0$ communicates with his neighbors once he has an interaction with POI $j_3$. }
\label{randomwalk}
\vskip -0.15in
\end{figure}

With the adjacency matrix $W$ representing the communication graph among users, \textit{how far should users communication with their neighbors} becomes the second challenging question. 
Practically, a user's decision on an item is not only affected by his direct neighbors, but also the further neighbors, e.g., the neighbors of neighbors. 
Thus, when a user $i$ rates an item $j$, this information should not only be sent to the directly neighbors of user $i$, but also his further neighbors, as shown in Figure \ref{randomwalk}. 
The challenge remains to be how far to explore the network, since there is a tradeoff between decentralization and communication/computation cost: 
the further the communication is, the more users can collaborate, meanwhile, the more communication and computation need to be done. 
We propose to solve this challenge by using random walk theory, which has been used to model trust relationship between users \cite{jamali2009trustwalker}.

\textbf{Random walk. }
We aim to find an intelligent way to determine how far a user communicate with his neighbors. 
Assuming user $i$ wants to communicate with his direct neighbors ($k \in \mathcal{N}^1(i)$), we define $n_i$ as the activity of user $i$ selecting a user from his neighbor set, and thus 
\begin{equation}\small
P(n_i = k) = \frac{w_{ik}}{\sum\limits_{i' \in \mathcal{N}^1(i)}{w_{ii'}}}. 
\end{equation} 
According to the Markov property \cite{aldous2002reversible}, the probability of user $i$ choosing his $2^{st}$ order of neighbors ($k' \in \mathcal{N}^2(i)$) is
\begin{equation}\small
P(n_i = k') = \sum\limits_{k} P(n_i = k)  P(n_k = k') \propto \sum\limits_{k} w_{ik} w_{kk'}. 
\end{equation} 
We use $D$ to denote the max distance of random walk, and generally, the adjacent matrix of $d^{st}$ order of neighbors is $W^d$. 
With random walk theory on user adjacent graph, the second question is answered.

\subsection{DMF: Privacy Preserving Nearby User Communication.} 
The random walk based nearby user communication is an intelligent way for selecting neighbors to be communicated within POI recommendation scenarios, but the third challenging question remains: \textit{what information should users communicate with each other}, that is, how should users collaboratively learn the DMF model without leaking their data privacy. 
The original rating, of course, can clearly reflect his preference on this item. 
However, the rating itself discloses the user's privacy too much. 
Inspired by the work \cite{yan2013distributed}, we propose a privacy preserving collaborative approach for decentralized POI recommendation scenarios. 
Specifically, we suppose that for each user, the corresponding $j$-th item latent factor $v^i_j$ can be decomposed as follows: 
\begin{equation}\label{decom1}
v^i_j=p_j+q^i_j,
\end{equation} 
which implies that the latent factor of item $j$ for user $i$ is the sum between \textit{one common (global) latent factor} $p_j$ and \textit{one personal (local) latent factor} $q^i_j$, where the common factor represents the common preference of all the users while the personal factor shows the personal favor of user $i$. 
Under this assumption, the DMF model can be formulated as 
\begin{equation}\small
\label{comdmf}
\begin{split}
	\mathop { \min }\limits_{u_i,v^i_j \in \mathbb{R}^K} \mathcal{L} 
	& = \sum\limits_{i=1}^{I} l(r,u_i,v^i) + \frac{\alpha}{2} \sum\limits_{i=1}^I  ||{u_i}||_F^2 \\
	& + \frac{\beta}{2} \sum\limits_{j=1}^J||p_j||_F^2 + \frac{\gamma}{2} \sum\limits_{i=1}^I \sum\limits_{j=1}^J||q^i_j||_F^2 \\
	& ~~{\rm s.t.} ~~ v^i_j=p_j+q^i_j,
\end{split}	
\end{equation} 
where $l(r,u_i,V^i)$ can be least square loss
\begin{equation}\small
l(r,u_i,V^i)=\frac{1}{2}\sum\limits_{j=1}^{J}{\left(r_{ij} - u_i^T v^i_j\right)}^2,
\end{equation} 
which minimizes the error between real ratings and predicted ratings \cite{mnih2007probabilistic}, or listwise loss \cite{shi2010list}, as well as pairwise loss \cite{rendle2009bpr}. 
In this paper, we will take least square loss as an example. 
The last three terms in Equation (\ref{comdmf}) are regularizers to prevent overfitting.

The factors $u_i$ and $q^i_j$ only depend on the information stored in user $i$, while the item common latent factor $p_j$ depends on the information of all the users.
Practically, in decentralized learning scenario, $p_j$ is also saved on each user's (learner's) hand, and thus, for each user $i$, $p_j$ is actually saved as $p^i_j$. Consequently, Equation (\ref{decom1}) becomes 
\begin{equation}\label{decom2}
v^i_j=p^i_j+q^i_j.
\end{equation} 
Thus, it needs one protocol for users to exchange $p^i_j$ to learn a global $p_j$. 
To solve this issue, inspired by \cite{yan2013distributed}, we propose to send the gradient of the loss $\mathcal{L}$ with respect to $p_j$, i.e., $p^i_j$ for each user $i$, to his neighbors, to help learn a global $p_j$.  
This gradient exchange method has been successfully applied in decentralized learning scenarios \cite{nedic2009distributed,yan2013distributed}, which not only guarantees model convergency, but also protects the privacy of user raw data.  
For each user $i$, the gradients of $\mathcal{L}$ with respect to $u_i$, $p^i_j$, and $q^i_j$ are:
\begin{equation}\small
\label{usergradient}
\frac{\partial \mathcal{L}}{\partial u_i} = -{\left(r_{ij} - u_i^T v^i_j\right)}v^i_j + \alpha u_{i},
\end{equation} 
\begin{equation}\small
\label{commonitemgradient}
\frac{\partial \mathcal{L}}{\partial p^i_j} = -{\left(r_{ij} - u_i^T v^i_j\right)}u_i + \beta p^i_j,
\end{equation} 
\begin{equation}\small
\label{personalitemgradient}
\frac{\partial \mathcal{L}}{\partial q^i_j} = -{\left(r_{ij} - u_i^T v^i_j\right)}u_i + \gamma q^i_j.
\end{equation}

Based on the above gradient exchange protocol, users collaboratively learn a global $p_j$. 
Figure \ref{randomwalk} shows a demo of this protocal: $i_0$ will send ${\partial \mathcal{L}}/{\partial p^0_3}$ to his neighbors, to collaboratively learn $p_3$. 
Combining our proposed random walk enhanced nearby user communication method and gradient exchange protocol, we summarize our proposed privacy preserving DMF optimization framework for POI recommendation (Equation \ref{comdmf}) in Algorithm \ref{dmflearning}. 
As we can see that users communicate with each other by sending the common item latent factor gradients instead of raw data, i.e., ratings, which significantly reduces the possibility of information leakage. 
With the gradient exchange protocol, the third question is answered.

From the objective function of DMF in Equation (\ref{comdmf}), we can easily make the following observations:
\begin{itemize}
\item If $\beta$ is very large, then $p_j \rightarrow 0$, and thus users will not exchange item common preferences, which means that item preference is learnt only based on their own data. 
\item If $\gamma$ is very large, then $q^i_j \rightarrow 0$, which indicates that users will not save their personal favor on items anymore. 
It will work more like centralized MF. 
\end{itemize}

The values of $\beta$ and $\gamma$ determine how well item (common and personal) preferences are learnt. We will empirically study their effects on our model performance in experiments. 

\begin{algorithm}[t]\label{dmflearning}
\caption{Random Walk Enhanced Nearby Collaborative DMF Optimization}
\KwIn {training ratings ($\mathcal{O}$), learning rate ($\theta$), user adjacency matrix ($W$), regularization strength($\alpha$, $\beta$, $\gamma$), maximum random walk distance ($D$), and maximum iterations ($T$)}
\KwOut{user latent factor ($u_i$), common item latent factor ($p^i_j$), and personal item latent factor ($q^i_j$)}

\For{$i=1$ to $I$}{
	Initialize $u_i$, $p^i_j$, and $q^i_j$
}
\For{$t=1$ to $T$}
{
	Shuffle training data $\mathcal{O}$\\
	\For{$r_{ij}$ in $\mathcal{O}$}{
		Calculate $\frac{\partial \mathcal{L}}{\partial u_i}$ based on Equation (\ref{usergradient}) \\
		Calculate $\frac{\partial \mathcal{L}}{\partial p^i_j}$ based on Equation (\ref{commonitemgradient}) \\
		Calculate $\frac{\partial \mathcal{L}}{\partial q^i_j}$ based on Equation (\ref{personalitemgradient})\\
		Update $u_i$ by $u_i \leftarrow u_i - \theta\frac{\partial \mathcal{L}}{\partial u_i}$ \\
		Update $p^i_j$ by $p^i_j \leftarrow p^i_j - \theta\frac{\partial \mathcal{L}}{\partial p^i_j}$ \\
		Update $q^i_j$ by $q^i_j \leftarrow q^i_j - \theta\frac{\partial \mathcal{L}}{\partial q^i_j}$\\
		\For{user $i'$ in $\mathcal{N}^d(i),d \in \{1,2,...,D\}$}{
			Receive $\frac{\partial \mathcal{L}}{\partial p^i_j}$ from user $i$ \\
			Update $p^{i'}_j$ by 
			$\small p^{i'}_j \leftarrow p^{i'}_j - \theta {|\mathcal{N}^d(i)|} W_{ii'}\frac{\partial \mathcal{L}}{\partial p^i_j}$
		}
	}
}
\Return $u_i$, $p^i_j$, and $q^i_j$
\end{algorithm}


\textbf{Unobserved rating sample.} 
A universal problem in POI recommendation is that the observations are extremely sparse. 
Unless we have access to negative observations, we will probably obtain an estimator that tends to predict all the unknown $(i,j)$ as 1. 
Following the existing researches \cite{yang2011like}, we solve this problem by sampling unobserved $(i,j)$ during SGD optimization. 
Specifically, for each $r_{ij} \in \mathcal{O}$, we randomly sample $m$ missing entries $r_{{ij'}, j'=1:m}$ and treat them as negative examples, i.e., $r_{ij'}=0$. 
However, a missing entry $r_{ij'}$ can denote either $i$ does not like $j'$ or $i$ does not know the existing of $j'$. 
Therefore, we decrease the confidence of $r_{ij'}$ to $1/m$. 



\subsection{Complexity Analysis}
Here we analyze the communication and computation complexity of Algorithm \ref{dmflearning}. 
Recall that $K$ denotes the dimension of latent factor, $N$ denotes the maximum number of direct neighbors of each user, $D$ denotes the max distance of random walk, $\mathcal{O}$ as the training data, and $|\mathcal{O}|$ as its size. 

\textbf{Communication Complexity. }
The communication cost depends on both the length of item gradient and number of neighbor to be communicated. 
Each item gradient contains $4K$ bytes information, since it is a $K$ dimensional real-valued vector. 
For user $i$, the max number of neighbors to be communicated of $D^{st}$ order random walk is $\min(|C^i|, \mathcal{N}^D(i))$, where $|C^i|$ is the number of users in the current city of $i$. 
Thus, for each $r_{ij} \in \mathcal{O}$, the communication cost is $\min(|C^i|, \mathcal{N}^D(i)) \cdot 4K$ bytes. 
It will be $|\mathcal{O}| \cdot \min(|C^i|, \mathcal{N}^D(i)) \cdot 4K$ bytes for passing the training dataset once. 
The values of $N$, $D$, and $K$ are usually small, and thus the communication cost is linear with the training data size. 

\textbf{Computation Complexity. }
The computation cost mainly relies on two parts, (1) calculating gradients, i.e., Line $7-9$ in Algorithm \ref{dmflearning}, and (2) updating user and item latent factors, i.e., Line $10-12,15$ in Algorithm \ref{dmflearning}. 
For a single pass of the training data, the time complexity of (1) is $|\mathcal{O}| \cdot K$, and the time complexity of (2) is $|\mathcal{O}| \cdot \min(|C^i|, \mathcal{N}^D(i)) \cdot K$. 
Therefore, the total computational complexity in one iteration is $|\mathcal{O}| \cdot \min(|C^i|, \mathcal{N}^D(i)) \cdot K$. 
The values of $N$, $D$, and $K$ are usually small, and thus the time complexity is linear with the training data size. 

The above communication and computation complexity analysis shows that our proposed approach is very efficient and can scale to very large datasets. 


%% file: section/experiment.tex
\section{Experiments}\label{experiments}
In this section, we empirically compare the performance of DMF with the classic centralized MF models, we also study the effects of parameters on model performance. 

\subsection{Setting}
We first describe the datasets, metrics, and comparison methods we use during our experiments. 

\textbf{Datasets.} We use two real-world datasets, i.e., \emph{Foursquare} and \emph{Alipay}. 
\emph{Foursquare} is a famous benchmark dataset for evaluating a POI recommendation model \cite{yang2016participatory}. 
We randomly choose two cities for each country from the original dataset, and further remove the users and POIs that have too few or too many interactions\footnote{The reason we sample a small dataset is: we mock decentralized learning during our experiments, that is, there will be $2I \times \mathbb{R}^{K\times{}J}$ POI (global and local) latent matrices in total, which actually is not a small scale.}. 
Our \emph{Alipay} dataset is sampled from user-merchant offline check-in records during 2017/07/01 to 2017/07/31, and we also perform similar preprocess on it. 
Table \ref{dataset} shows the statistics after process for both datasets, with which we randomly sample 90\% as training set and the rest 10\% as test set. 

\textbf{Metrics.} 
We adopt two metrics to evaluate our model performance, i.e., $P@k$ and $R@k$, which are commonly used to evaluate POI recommendation performance \cite{cheng2012fused,gao2015content}. 
For user $i$, they are defined as
\begin{equation}\nonumber\small
	P@k=\frac{|S^T_i \cap S^R_i|}{k},~~~
	R@k=\frac{|S^T_i \cap S^R_i|}{|S^T_i|},
\end{equation} 
where $S^T_i$ denotes the visited POI set of user $i$ in the test data, and $S^R_i$ denotes the recommended POI set of user $i$ which contains $k$ POIs.

\begin{table}
\centering
\caption{Dataset statistics.}
\label{dataset}
\begin{tabular}{|c|c|c|c|c|}
  \hline
  Dataset & \#user & \#item & \#rating & \#cities \\
  \hline
  \hline
  \emph{Foursquare} & 6,524 & 3,197 & 26,186 & 117 \\
  \hline
  \emph{Alipay} & 5,996 & 7,404 & 18,978 & 298 \\
  \hline
\end{tabular}
\end{table}

\textbf{Comparison methods.} 
Our proposed DMF framework is a novel decentralized algorithm for POI recommendation, which is a decentralized version for the classic MF model \cite{mnih2007probabilistic}. 
We compare our proposed DMF with the following classic and state-of-the-art latent factor models, including several variants of DMF:

\begin{itemize}[leftmargin=*] \setlength{\itemsep}{-\itemsep}
    \item \textbf{MF} \cite{mnih2007probabilistic} is a classic centralized latent factor model which uses least square loss. 
    \item \textbf{Bayesian Personalized Ranking (BPR)} \cite{rendle2009bpr} is the state-of-the-art centralized latent factor model which uses pairwise loss. 
    \item \textbf{Global DMF (GDMF)}. Users will not save their personal favor ($q^i_j$) anymore, and they tend to share similar latent factor for the same item. This is a special case of our proposed DMF model, i.e., when $\gamma$ is very large. 
    \item \textbf{Local DMF (LDMF)}. Users will not exchange preferences and they learn the model only based on their own data. This is also a special case of our proposed DMF model, i.e., when $\beta$ is very large. 
\end{itemize}

Note that we do not compare with the state-of-the-art POI recommendation methods. This is because, (1) most of them are the improvement of the classic MF model by using additional information, e.g. user social information and contextual information \cite{cheng2012fused,yang2013sentiment,gao2015content}, which are not fair to compare with, and 
(2) our focus is to compare the effectiveness of the traditional centralized latent factor models and our proposed decentralized MF model. 

\textbf{Hyper-parameters.} 
During comparison, we set user regularizer $\alpha=0.1$, learning rate $\theta=0.1$, and the returned number of POI $k \in \{5,10\}$. 
We also set the maximum number of neighbor $N=2$, and the number of sampled unobserved ratings $m=3$. 
After we build the user adjacent graph, we simply set $w_{i,i'}=1$ to eliminate the effect of mapping function on model performance, since this is not the focus of this paper. 
For the latent factor dimension $K$, we vary its values in $\{5,10,15\}$. 
For the random walk distance $D$, we vary its values in $\{1,2,3,4\}$. 
We also vary $\beta$ and $\gamma$ in $\{10^{-3},10^{-2},10^{-1},10^{0},10^{1}\}$ to study their effects on DMF. 
We tune parameters of each model to achieve their best performance for comparison. 

\begin{table}[t]
\centering
\caption{Performance comparison on \emph{Foursquare} dataset.}
\label{Foursquare}
\begin{tabular}{|c|p{1.15cm}<{\centering}|p{1.15cm}<{\centering}|p{1.15cm}<{\centering}|p{1.15cm}<{\centering}|}
 \hline
   Metrics & P@5 & R@5 & P@10 & R@10 \\
  \hline
  \hline
  Dimension & \multicolumn{4}{c|}{\emph{K=5}} \\
  \hline
  MF & 0.0291 & 0.1030 & 0.0242 & 0.1697 \\
  \hline
  BPR & 0.0293 & 0.1168 & 0.0247 & 0.2058 \\
  \hline
  GDMF & 0.0307 & 0.1245 & 0.0263 & 0.2115 \\
  \hline
  LDMF & 0.0025 & 0.0125 & 0.0025 & 0.0125 \\
  \hline
  DMF & \textbf{0.0337} & \textbf{0.1562} & \textbf{0.0291} & \textbf{0.2418} \\
  \hline
  \hline
  Dimension & \multicolumn{4}{c|}{\emph{K=10}} \\
  \hline
  MF & 0.0289 & 0.1250 & 0.0263 & 0.2226 \\
  \hline
  BPR & 0.0370 & 0.1533 & 0.0302 & 0.2514 \\
  \hline
  GDMF & 0.0281 & 0.1286 & 0.0269 & 0.2398 \\
  \hline
  LDMF & 0.0133 & 0.0565 & 0.0133 & 0.1121 \\
  \hline
  DMF & \textbf{0.0386} & \textbf{0.1553} & \textbf{0.0340} & \textbf{0.2774} \\
  \hline
  \hline
  Dimension & \multicolumn{4}{c|}{\emph{K=15}} \\
  \hline
  MF & 0.0406 & 0.1711 & 0.0316 & 0.2638 \\
  \hline
  BPR & \textbf{0.0448} & \textbf{0.1898} & 0.0342 & 0.2872 \\
  \hline
  GDMF & 0.0316 & 0.1388 & 0.0287 & 0.2500  \\
  \hline
  LDMF & 0.0187 & 0.0714 & 0.0170 & 0.1511 \\
  \hline
  DMF & 0.0421 & 0.1801 & \textbf{0.0370} & \textbf{0.3035} \\
  \hline
\end{tabular}
\end{table}

\subsection{Comparison Results}

\begin{figure*}[t]
\centering
\subfigure [\emph{Foursquare} training loss]{ \includegraphics[width=4cm,height=2.35cm]{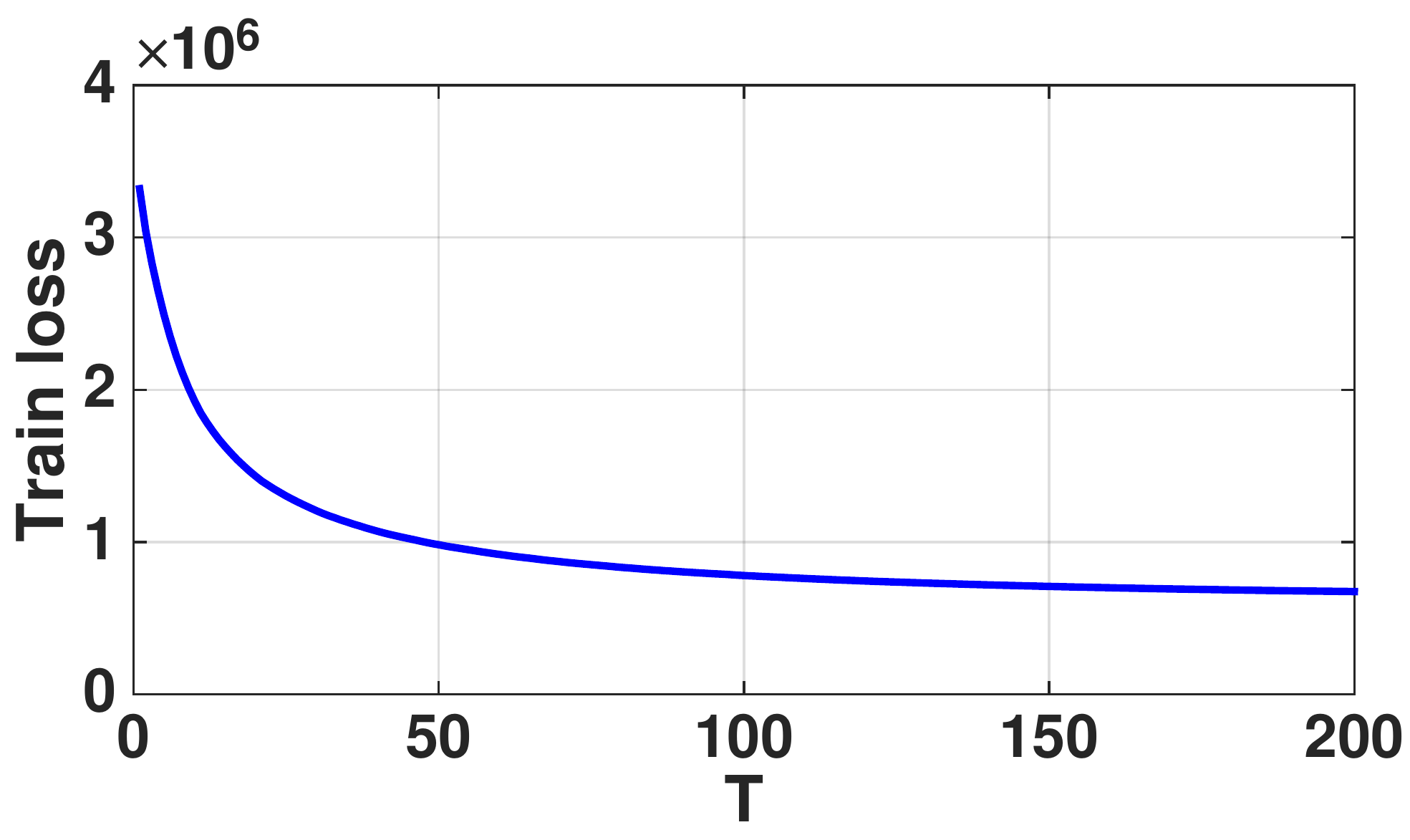}}~~~
\subfigure [\emph{Foursquare} test loss]{ \includegraphics[width=4cm,height=2.35cm]{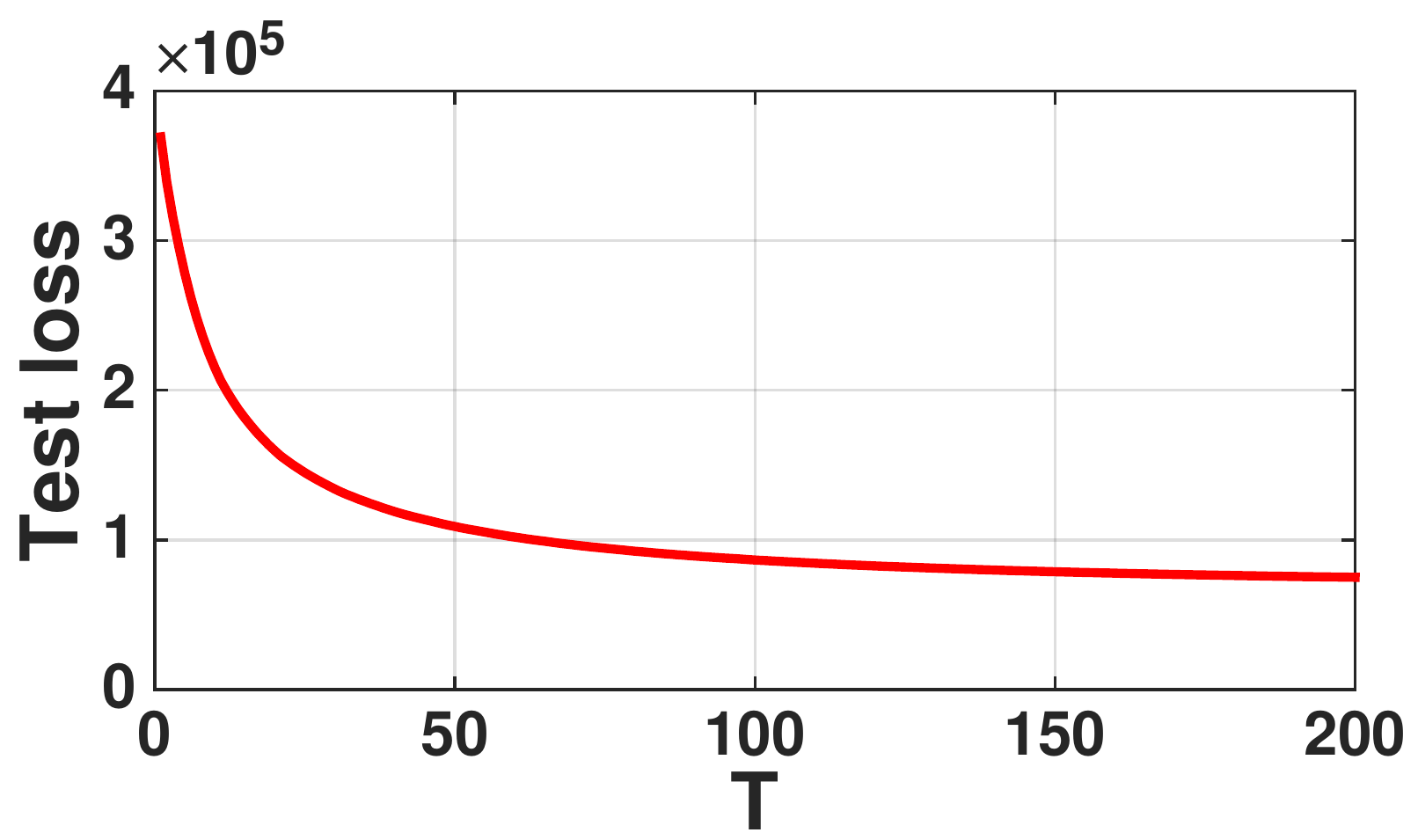}}~~~
\subfigure [\emph{Alipay} training loss]{ \includegraphics[width=4cm,height=2.35cm]{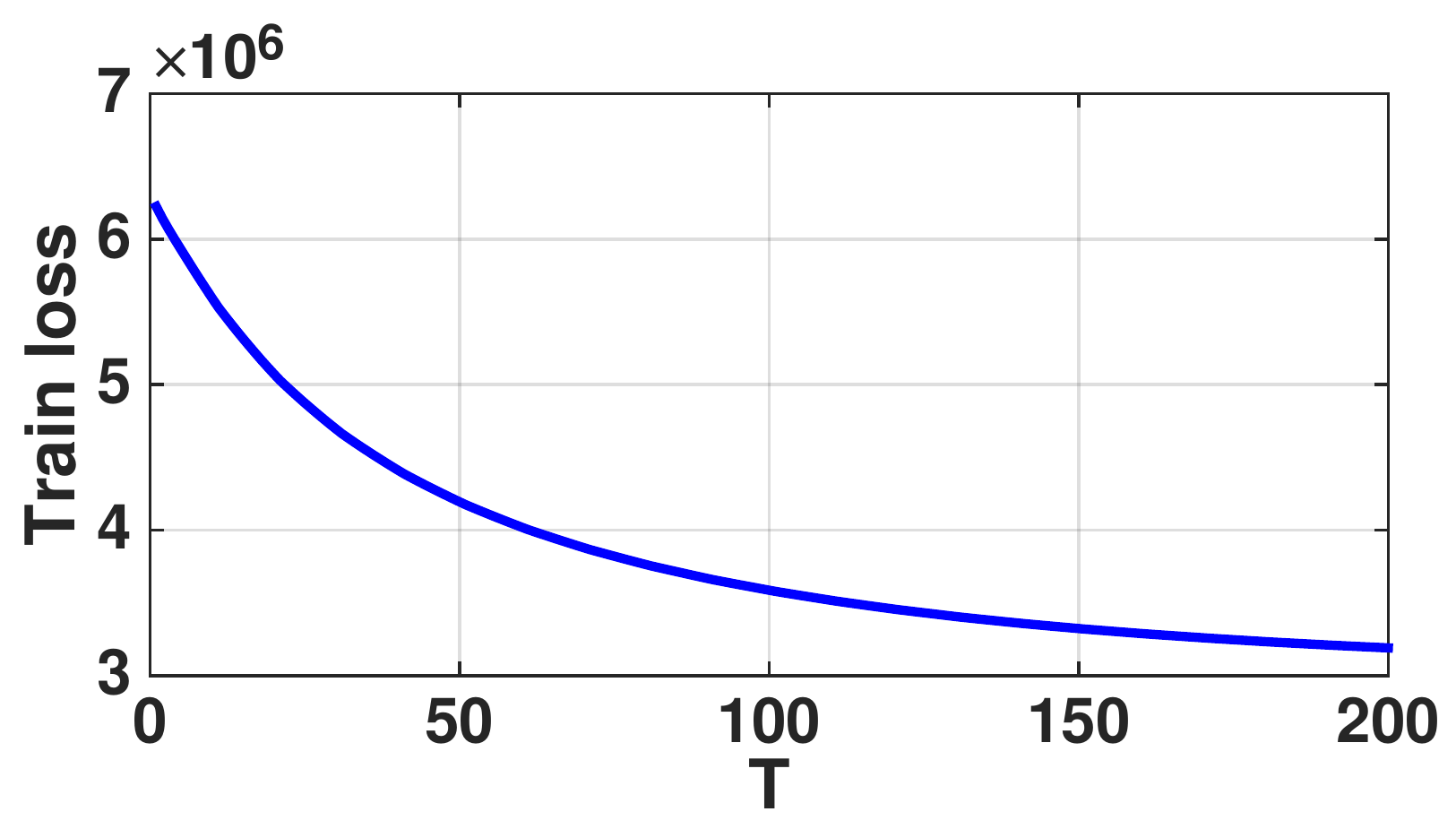}}~~~
\subfigure[\emph{Alipay} test loss] { \includegraphics[width=4cm,height=2.35cm]{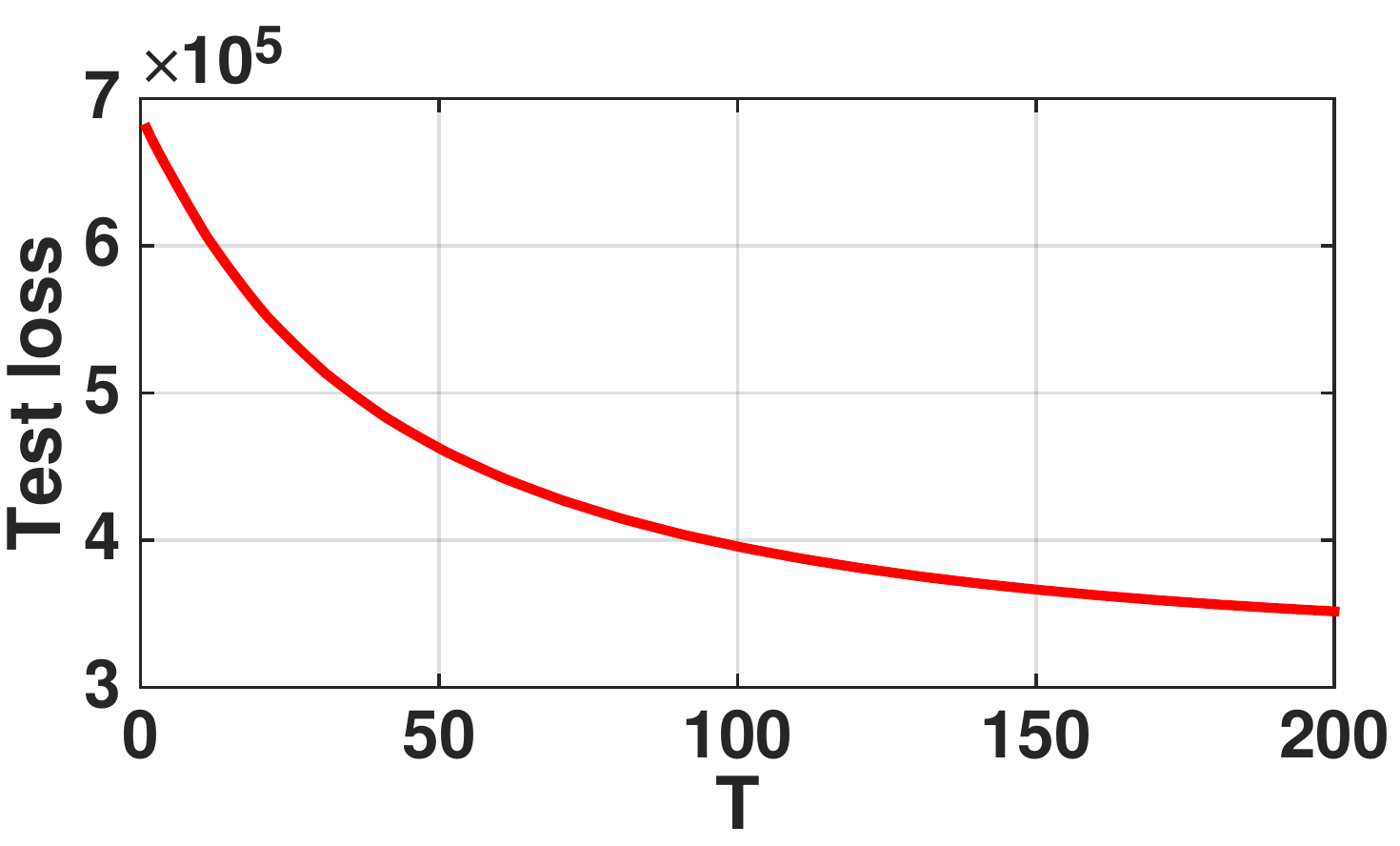}}
\vskip -0.1in
\caption{The training and test loss of DMF with respect to the maximum iteration ($T$) on \emph{Foursquare} and \emph{Alipay} datasets. }
\label{effectoft}
\end{figure*}

\begin{table}
\centering
\caption{Performance comparison on \emph{Alipay} dataset.}
\label{Alipay}
\begin{tabular}{|c|p{1.15cm}<{\centering}|p{1.15cm}<{\centering}|p{1.15cm}<{\centering}|p{1.15cm}<{\centering}|}
 \hline
   Metrics & P@5 & R@5 & P@10 & R@10 \\
  \hline
  \hline
  Dimension & \multicolumn{4}{c|}{\emph{K=5}} \\
  \hline
  MF & 0.0209 & 0.0896 & 0.0134 & 0.1194 \\
  \hline
  BPR & 0.0243 & 0.1027 & 0.0166 & 0.1408 \\
  \hline
  GDMF & 0.0209 & 0.0937 & 0.0144 & 0.1231 \\
  \hline
  LDMF & 0.0017 & 0.0042 & 0.0008 & 0.0042 \\
  \hline
  DMF & \textbf{0.0267} & \textbf{0.1128} & \textbf{0.0228} & \textbf{0.1824} \\
  \hline
  \hline
  Dimension & \multicolumn{4}{c|}{\emph{K=10}} \\
  \hline
  MF & 0.0343 & 0.1493 & 0.0261 & 0.2239 \\
  \hline
  BPR & 0.0353 & 0.1512 & 0.0286 & 0.2439 \\
  \hline
  GDMF & 0.0304 & 0.1146 & 0.0277 & 0.2202 \\
  \hline
  LDMF & 0.0013 & 0.0065 & 0.0013 & 0.0131 \\
  \hline
  DMF & \textbf{0.0357} & \textbf{0.1612} & \textbf{0.0291} & \textbf{0.2537} \\
  \hline
  \hline
  Dimension & \multicolumn{4}{c|}{\emph{K=15}} \\
  \hline
  MF & 0.0378 & 0.1538 & 0.0266 & 0.2308 \\
  \hline
  BPR & \textbf{0.0448} & \textbf{0.1898} & 0.0342 & 0.2872 \\
  \hline
  GDMF & 0.0355 & 0.1383 & 0.0267 & 0.2123 \\
  \hline
  LDMF & 0.0137 & 0.0684 & 0.0103 & 0.0983 \\
  \hline
  DMF & 0.0413 & 0.1942 & \textbf{0.0370} & \textbf{0.3035} \\
  \hline
\end{tabular}
\end{table}

We report the comparison results on both \emph{Foursquare} and \emph{Alipay} datasets in Table \ref{Foursquare} and Table \ref{Alipay}, respectively. 
From them, we find that:
\begin{itemize}[leftmargin=*] \setlength{\itemsep}{-\itemsep}
    \item All the model performances increase with the dimension of latent factor ($K$). 
	This is because the latent factor with a bigger $K$ contains more information, and thus user and POI preferences can be learnt more precisely. 
	\item GDMF achieves comparable performance with MF, since users collaboratively learn the global item latent factor, which works similarly as the traditional MF. 
	\item LDMF behave the worst, since each user learn item preference only based on his own check-in data which is very sparse. This indicates the importance of user collaboration during recommendation. 
    \item DMF consistently outperforms MF, GDMF, and LDMF, and even beats the pairwise ranking model (BPR) in most cases. 
    Take the result of DMF on \emph{Alipay} dataset for example, $P@5$ and $R@5$ of DMF improve those of MF by 27.75\% and 25.89\% when $K=5$. 
    This is because, the item preference of DMF contains not only common preference that obtained from all the users' data, but also each user's  personal preference that learnt from his own data. 
    Moreover, the random walk technique intelligently help users to choose neighbors to communicate with. 
    Therefore, item preferences are learnt more precisely, to better match each user's favor. 
\end{itemize}

\subsection{Parameter Analysis}
Finally, we study the effects of parameters on DMF, including item global and local regularizer ($\beta$ and $\gamma$), maximum random walk distance ($D$), and maximum iteration ($T$).

\textbf{Effect of $\beta$ and $\gamma$.} 

\begin{figure}[t]
\centering
\subfigure [\emph{Foursquare}]{ \includegraphics[width=4cm]{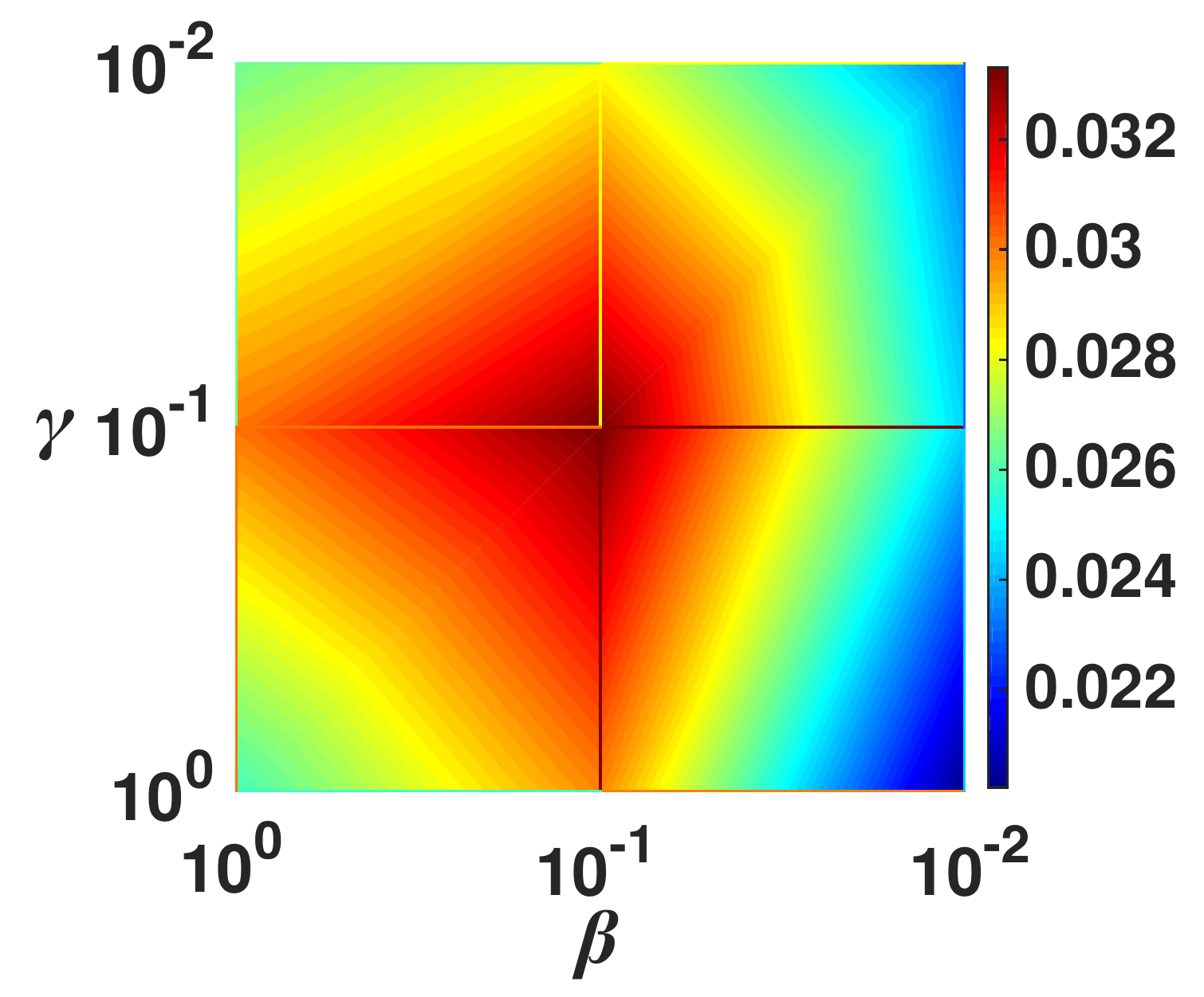}}~~~
\subfigure[\emph{Alipay}] { \includegraphics[width=4cm]{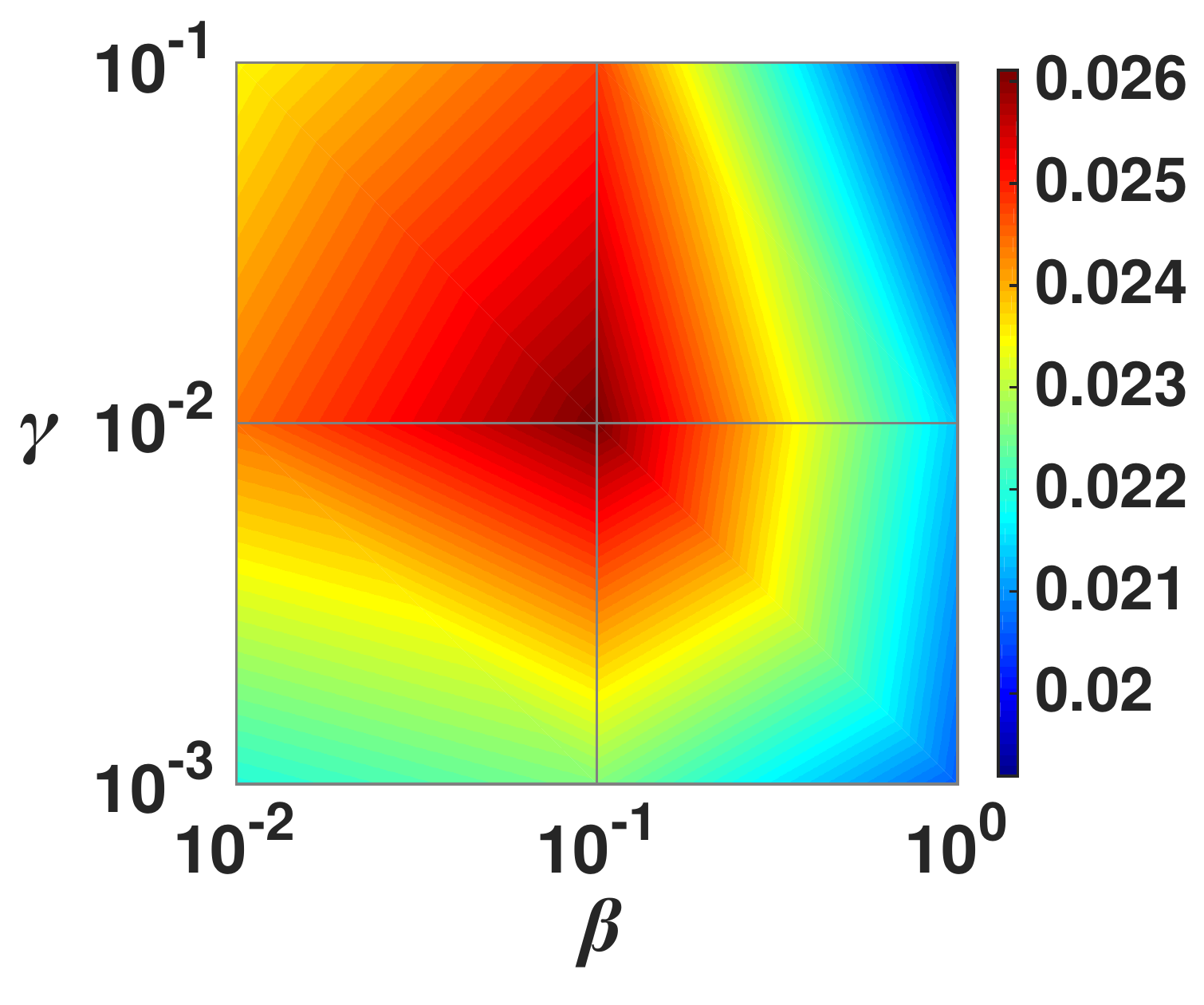}}
\vskip -0.05in
\caption{Effect of $\beta$ and $\gamma$ on DMF. }
\label{effectofl}
\end{figure}

The item global regularizer ($\beta$) and local regularizer ($\gamma$) controls the proportion of one's item preference comes from his own data or other users' data. 
The bigger $\beta$ is, the more one's item preference comes from his own data, and similarly, the bigger $\gamma$ is, the more one's item preference comes from other users' data through global item gradient (${\partial \mathcal{L}}/{\partial p^i_j}$) exchange. 
Figure \ref{effectofl} shows their effects on DMF on both datasets. 
From it, we can find that, with the good choices of $\beta$ and $\gamma$, DMF can make full use of one's own data and his neighbors' data, and thus, achieves the best performance. 

\begin{figure}
\centering
\subfigure [\emph{Foursquare}]{ \includegraphics[width=4cm]{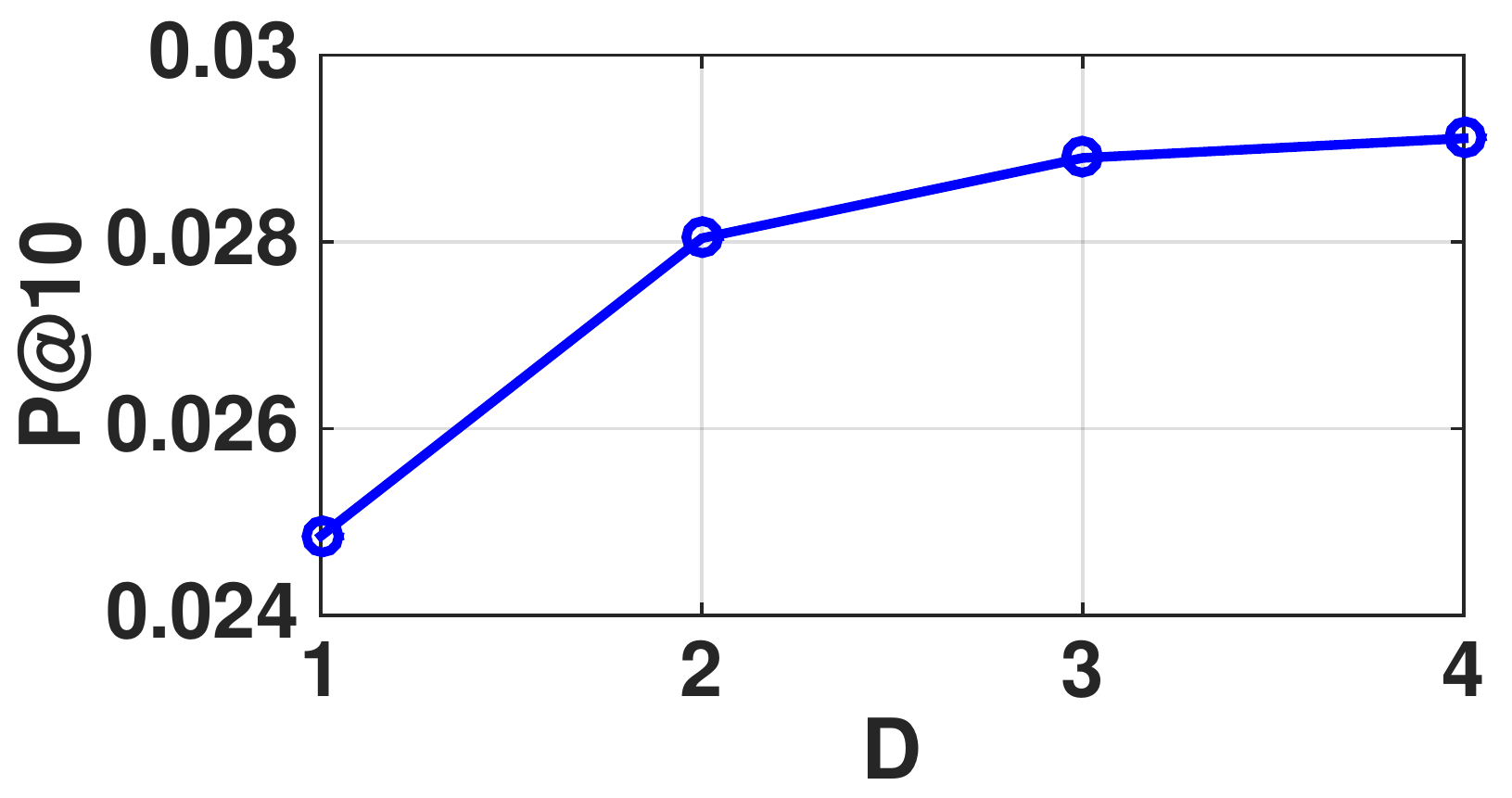}}~~~
\subfigure[\emph{Alipay}] { \includegraphics[width=4cm]{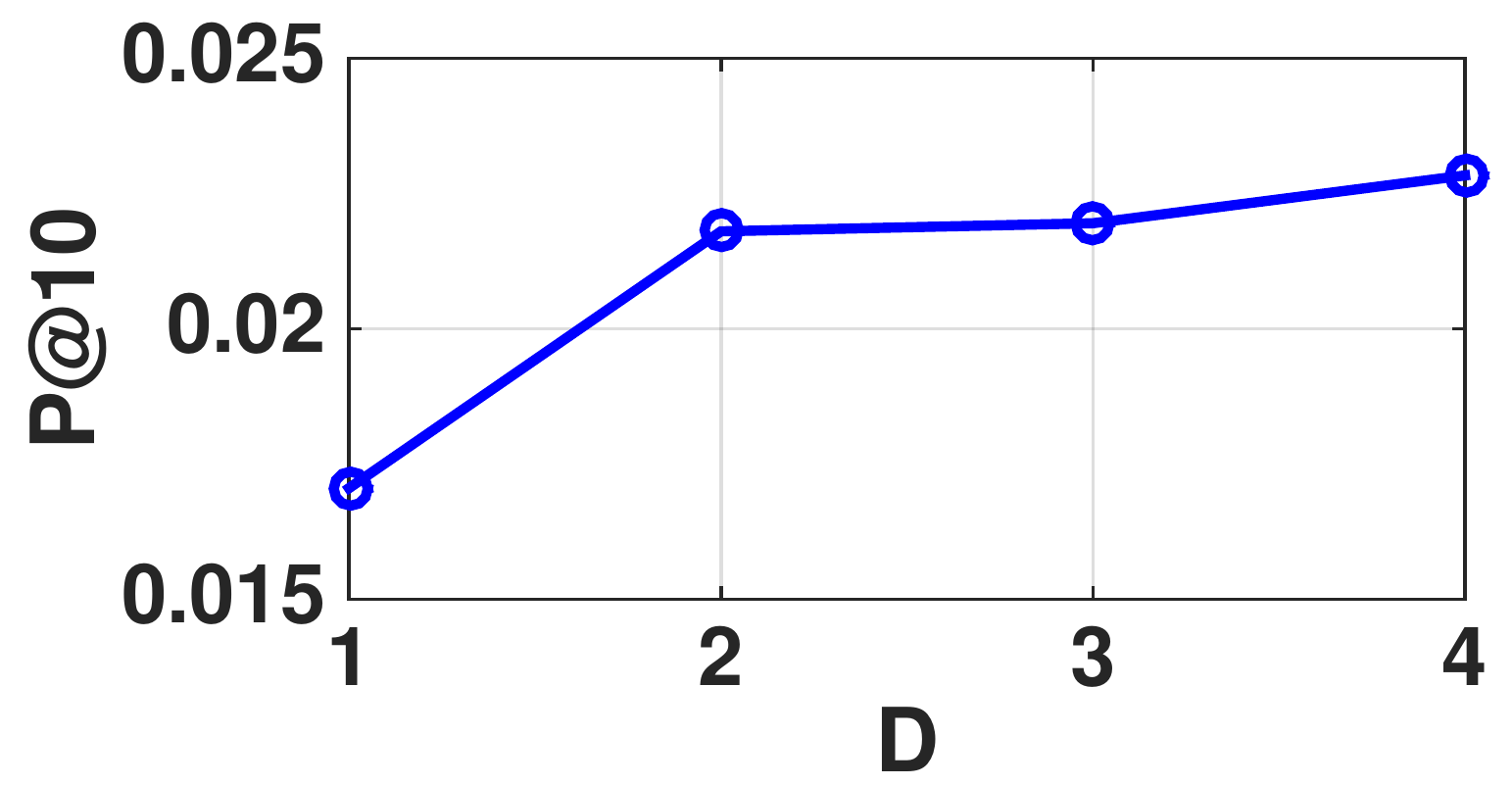}}
\vskip -0.05in
\caption{Effect of $D$ on DMF. }
\label{effectofd}
\vskip -0.15in
\end{figure}

\textbf{Effect of maximum random walk distance ($D$).} 
The maximum random walk distance determinates how many neighbors will be communicated after a user has interaction with a POI, as we described in complexity analysis section. 
Figure \ref{effectofd} shows its effect on DMF model on both \emph{Foursquare} and \emph{Alipay} datasets, where we set $K=5$ and fix other parameters to their best values. 
From it, we see that with the increase of $D$, our model performance increases, and tends to be relative stable when $D$ is bigger than 3. 
This shows that DMF achieves a good performance with only a small value of $D$, which indicates a low cost of communication complexity. 

\textbf{Effect of maximum iteration ($T$).} 
As we analyzed above, the computing time complexity is linear with the training data size, and thus, the converge speed determines how long DMF should be trained. 
Figure \ref{effectoft} shows the effect of $T$ on training loss and test loss on both \emph{Foursquare} and \emph{Alipay} datasets. 
As we can observe, DMF converges steadily with the increase of $T$, and it takes about 100 epochs to converge on \emph{Foursquare} and about 200 epochs on \emph{Alipay}.

%% file: section/conclusion.tex
\section{Conclusion and Future Work}

In this paper, we proposed a Decentralized MF (DMF) framework for POI recommendation. 
Specifically, we proposed a random walk based nearby collaborative decentralized training technique to train DMF model in each user's end. 
By doing this, the data of each user on items are still kept on one's own hand, and moreover, decentralized learning can be taken as distributed learning with multi-learner (user), and thus solves the efficiency problem. 
Experimental results on two real-world datasets demonstrate that, comparing with the classic and state-of-the-art latent factor models, DMF significantly improvements the recommendation performance in terms of precision and recall. 

We would like to take model compression as our future work. 
Currently, each user needs to store the real-valued item latent matrix. 
A binary type of latent matrix will significantly reduce the storage cost. 
How to balance the model storage and model accuracy will be our next stage of work.